\documentclass[12pt,twoside]{article}   %For printing on two sides of page
\usepackage[super,sort,comma]{natbib}
\usepackage{amsfonts}
\usepackage{amsmath}
\usepackage{amssymb}
\usepackage{mathrsfs}
\usepackage{amsbsy}
\usepackage{tensor}
\usepackage{esint}
\usepackage{tikz} 
\usepackage{multirow}
\usepackage{tabularx}
\usepackage{adjustbox}
\usepackage{makecell}
\usepackage{textcomp}
\usepackage{theorem}
\usepackage{epstopdf}
\usepackage[normalem]{ulem}
\usepackage{booktabs,floatrow}

\usepackage{fancyhdr}		%Gives headers and footers defined below
\usepackage[section]{placeins}   %
\usepackage{graphicx}

\makeatletter \renewcommand\@biblabel[1]{$^{#1}$} \makeatother
\newcommand{\cen}[1]{\begin{center} #1 \end{center}}

\usepackage[mathlines]{lineno}
\usepackage{hyperref}
\hypersetup{ colorlinks,
    citecolor=blue,
    filecolor=blue,
    linkcolor=blue,
    urlcolor=blue
}

\usepackage{xcolor}
\definecolor{gray}{rgb}{0.6,0.6,0.6}
\definecolor{red}{rgb}{0.85,0,0}
\definecolor{green}{rgb}{0,0.85,0}
\definecolor{blue}{rgb}{0,0,0.85}
\definecolor{beige}{rgb}{0.92,0.87,0.78}
\usepackage[all]{hypcap}    %causes link to figures to go to figure, not caption
                    {$\blacksquare$\vspace*{7pt}} % square <--> blacksquare

\def\0{{\mathbf{0}}}

\newcolumntype{C}{>{\centering\arraybackslash}p{7.5em}}

\begin{document}

\cen{\sf {\Large {\bfseries An Iterative Reconstruction Method for Dental Cone-Beam Computed Tomography with a Truncated Field of View} \\
\vspace*{10mm}
Hyoung Suk Park and Kiwan Jeon} \\
National Institute for Mathematical Sciences, Daejeon, 34047, Republic of Korea\\
% $^2$School of Mathematics and Computing (Computational Science and Engineering), Yonsei University, Seoul, 03722, Republic of Korea
%$^*$Corresponding author: jeonkiwan@nims.re.kr
%\vspace{5mm}\\
%Version typeset \today\\
}

\pagenumbering{roman}
\setcounter{page}{1}
\pagestyle{plain}
% Correspondence: Kiwan Jeon Email: jeonkiwan@nims.re.kr \\
% note, probably best not to use a student's e-mail as it won't be valid for
% very long.

\begin{abstract}
In dental cone-beam computed tomography (CBCT), compact and cost-effective system designs often use small detectors, resulting in a truncated field of view (FOV) that does not fully encompass the patient's head. In iterative reconstruction approaches, the discrepancy between the actual projection and the forward projection within the truncated FOV accumulates over iterations, leading to significant degradation in the reconstructed image quality. In this study, we propose a two-stage approach to mitigate truncation artifacts in dental CBCT. In the first stage, we employ Implicit Neural Representation (INR), leveraging its superior representation power, to generate a prior image over an extended region so that its forward projection fully covers the patient's head. To reduce computational and memory burdens, INR reconstruction is performed with a coarse voxel size. The forward projection of this prior image is then used to estimate the discrepancy due to truncated FOV in the measured projection data. In the second stage, the discrepancy-corrected projection data is utilized in a conventional iterative reconstruction process within the truncated region. Our numerical results demonstrate that the proposed two-grid approach effectively suppresses truncation artifacts, leading to improved CBCT image quality.
\end{abstract}

\newpage     %may or may not be needed

%The table of contents is for drafting and refereeing purposes only. Note
%that all links to references, tables and figures can be clicked on and
%returned to calling point using cmd[ on a Mac using Preview or some
%equivalent on PCs (see View - go to on whatever reader).
% \tableofcontents

\newpage

\setlength{\baselineskip}{0.7cm}      %double spacing		

\pagenumbering{arabic}
\setcounter{page}{1}
\pagestyle{fancy}
\section{Introduction}
Dental cone-beam computed tomography (CBCT) has gained popularity as a cost-effective, low-radiation alternative to multi-detector CT (MDCT) in dental clinics \cite{SCARFE2008,LUDLOW2008,KAASALAINEN2021}. It is widely used in various dental applications, including implant planning, orthodontic assessment, and orthognathic surgical planning \cite{Elnagar2020}. Typically, dental CBCT systems employ small flat-panel detectors with slower scanning speeds, enabling a lower cost and a compact design that requires significantly less space than MDCT systems. However, this design often results in a truncated field of view (FOV) that does not fully encompass the patient's head \cite{Park2024} and missing projection data, making the corresponding inverse problem ill-posed. 

Over the past few decades, various iterative reconstruction methods with suitable regularizations, such as total variation \cite{Tian2011}, wavelets \cite{Borsdorf2008}, dictionary learning \cite{Xu2012}, or learned and predefined regularizers \cite{Sun2019}, have been proposed for standard CT image reconstruction to enhance image quality. However, their application to dental CBCT under FOV truncation presents significant challenges. In iterative reconstruction approaches, the discrepancy between the actual projection and the forward projection within the truncated FOV accumulates over iterations, leading to significant degradation in the reconstructed image quality (see second column of Figure \ref{result_comparison}). 

To address this issue, Dang {\it et al}. \cite{DANG2017} proposed a multi-resolution iterative reconstruction method with an extended reconstruction FOV to mitigate truncation artifacts caused primarily by the head holder in their prototype CBCT system. Although their method demonstrated promising results in reducing truncation artifacts, its direct application to dental CBCT systems, which typically use smaller detectors and thus suffer from substantial missing projection data, may have limited performance.

Recently, Implicit Neural Representation (INR) has emerged as a powerful alternative for CT image reconstruction, representing images as continuous functions parameterized by neural networks. INR-based reconstruction efficiently captures complex spatial relationships between image pixels or voxels, significantly reduces the dimensionality of the solution space, and thereby enables more accurate and efficient reconstruction from highly undersampled data \cite{Park2025}. Several studies have applied INR to sparse-view CT image reconstruction \cite{Kim2022,Shen2022,ZHA2022}. Despite its considerable potential, computational complexity remain practical challenges in utilizing INR-based methods.

In this study, we propose a two-stage iterative reconstruction approach to reduce truncation artifacts in dental CBCT imaging. In the first stage, we utilize INR, taking advantage of its superior representation power, to reconstruct a prior image with a coarse voxel size over an extended region. The forward projection of this prior image fully covers the patient's head, enabling estimation and correction of discrepancies due to projection truncation. In the second stage, the corrected projection data is incorporated into a conventional iterative reconstruction process within the truncated region. Numerical evaluations demonstrate that the proposed two-grid strategy effectively mitigates truncation artifacts, leading to improved image quality in dental CBCT.

\begin{figure}[!ht]
\centering
\includegraphics[width=1.0\textwidth]{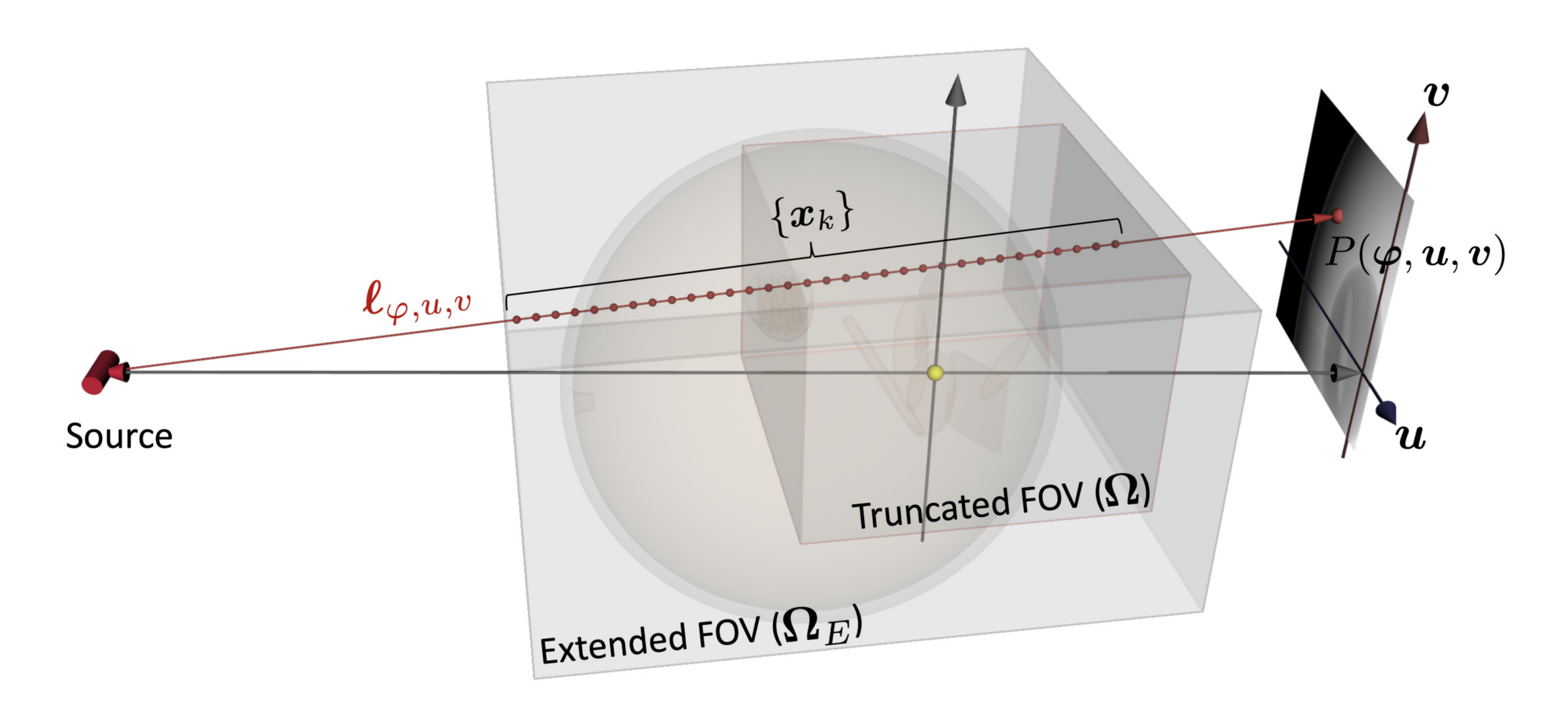}
\caption{Illustration of the 3D dental CBCT geometry with a truncated FOV.}
\label{geometry}
\end{figure}

Let $\Omega$ denote a truncated CBCT ROI caused by the compact system design with a small detector, and let $\Omega_E$ represent an extended region fully encompassing the patient’s head. The proposed method is a two-stage iterative reconstruction approach designed to mitigate truncation artifacts in dental CBCT imaging. In the first stage, we employ INR, leveraging its superior representation power, to reconstruct a prior image over the extended region $\Omega_E$ at a coarse voxel resolution (i.e., coarser than that used in the dental CBCT reconstruction. See Table \ref{tab_imaging_domain}). Specifically, for the 3D position ${\mathbf x}\in \Omega_E$, we first utilize a hash-grid-based positional encoding \cite{MULLER2022,ZHA2022}, denoted as $\text{ENC}(\mathbf{x})$, to efficiently represent spatial locations. These positional encodings serve as inputs to a multi-layer perceptron, denoted by $\text{MLP}_{\theta}(\cdot)$, which outputs the estimated attenuation values at each input position. According to the Lambert-Beer law \cite{Beer1852,Lambert1892}, the network parameters $\theta$ are optimized by minimizing the objective function ${\mathcal L}_c$ for the coarse-grid INR reconstruction, which is defined as:
\begin{equation}\label{inr_coarse}
   {\mathcal L}_c = \sum_{l_{\varphi,u,v}} \left|\text{P}(\varphi, u,v)- \sum_{{\mathbf x}_k\in l_{\varphi,u,v}}\text{MLP}_{\theta}(\text{ENC}({\mathbf x}_k))\,d_{{\mathbf x}_k} \right|,
\end{equation}
where, $\text{P}(\varphi,u,v)$ denotes the measured projection data at detector position $(u,v)$ and projection angle $\varphi$, $l_{\varphi,u,v}$ denotes the cone-beam ray connecting the X-ray source and the detector position $(u,v)$ at angle $\varphi$, ${\mathbf{x}}_k$ denotes sampled positions along the ray $l_{\varphi,u,v}$, and $d_{{\mathbf{x}}_k} = \|\mathbf{x}_{k+1}-\mathbf{x}_k\|$ denotes the distance between adjacent sample positions. The detailed geometry of the 3D dental CBCT system is illustrated in Figure \ref{geometry}.

From the trained network $\text{MLP}_{\theta}$, we generate the prior image ${\mathbf u}_{\text{o}}$ at the coarse grid positions $\mathbf{x}$ in $\Omega_E$, where all values inside $\Omega$ are set to zero. Using this prior image, the measured projection $\text{P}$ is adjusted to form the projection $\hat{\text{P}}$ for iterative reconstruction within the truncated region $\Omega$ as:
\begin{equation}\label{corrected_projection}
   \hat{\text{P}} = \text{P} - {\mathbf A}\,{\mathbf u_{\text{o}}},
\end{equation}
where ${\mathbf A}$ denotes the forward projection operator defined under the standard CBCT imaging geometry (i.e., truncated $\Omega$ and fine resolution).

In the second stage, the modified projection data is incorporated into a conventional iterative reconstruction framework:
\begin{equation}\label{iterative_fine}
    \arg\min_{{\mathbf u}}\frac{1}{2}\|{\mathbf A} {\mathbf u} - \hat{\text{P}} \|^2 + \lambda\,{\mathcal R}({\mathbf u}),
\end{equation}
where ${\mathcal R}(\cdot)$ represents a regularization term, and $\lambda$ is the corresponding regularization parameter.

% This approach enables efficient and continuous representation of attenuation coefficients while optimizing memory usage and computational efficiency.

\subsection{Implementation}
In order to provide both qualitative and quantitative verification, we use the Forbild head phantom~\cite{FORBILD}. The dental CBCT geometry used to generate truncated projection data is shown in the Table~\ref{tab_ct_geometry}.
\begin{table}[h]
    \centering
    \begin{tabular}{l c}
        % \hline
        \textbf{Property} & \textbf{Value} \\
        % \hhline{===}
        \hline\hline
        Distance (Source to Detector) & 600.0 \\
        Distance (Source and Iso Center) & 400.0 \\
        Number of Detector Pixels & $640 \times 640$ \\
        Detector Pixel Resolution  & $0.2 \times 0.2$ \\
        Detector Offset & (57.0, 29.0) \\
        Number of Projections & 300 \\
        Range of Projection Angle & 0\,--\,2$\pi$ \\
        \hline
    \end{tabular}
    \caption{CT geometry parameters for truncated projection data generation. The length unit is $mm$.}
    \label{tab_ct_geometry}
\end{table}

The imaging domain configuration for the coarse and fine grids is described in the Table~\ref{tab_imaging_domain}.

\begin{table}[h]
    \centering
    \begin{tabular}{l l c}
        % \hline
        \textbf{} & \textbf{Property} & \textbf{Values} \\ 
        \hline\hline
        % \hhline{===}
        \multirow{3}{*}{Coarse Grid} & Min. Position  & $\left[-140.0,\, -80.0,\, -40.0\right]$ \\
                                     & Max. Position  & $\left[ 140.0,\, 200.0,\, 105.0\right]$ \\
                                     & Voxel Resolution & $\left[ 1.0,\, 1.0,\,  1.0\right]$ \\
        \hline
        \multirow{3}{*}{Fine Grid}   & Min. Position  & $\left[-80.0,\, -80.0,\, -32.0\right]$ \\
                                     & Max. Position  & $\left[80.0,\, 80.0,\,  88.0\right]$ \\
                                     & Voxel Resolution & $\left[0.2,\, 0.2,\, 0.2\right]$ \\
        \hline
    \end{tabular}
    \caption{Imaging domain~(bounding box) description for the coarse and fine grid. The unit is $mm$.}
    \label{tab_imaging_domain}
\end{table}

For the implementation of INR, we used the PyTorch framework~\cite{PYTORCH}. The forward and backward projections required for iterative reconstruction were performed using the Astra Toolbox~\cite{ASTRA}. We minimize the objective function (\ref{iterative_fine}) using the separable quadratic surrogate~(SQS) method~\cite{ELBAKRI2002}.

\section{Results}

\begin{figure*}[!ht]
\centering
\includegraphics[width=1.0\textwidth]{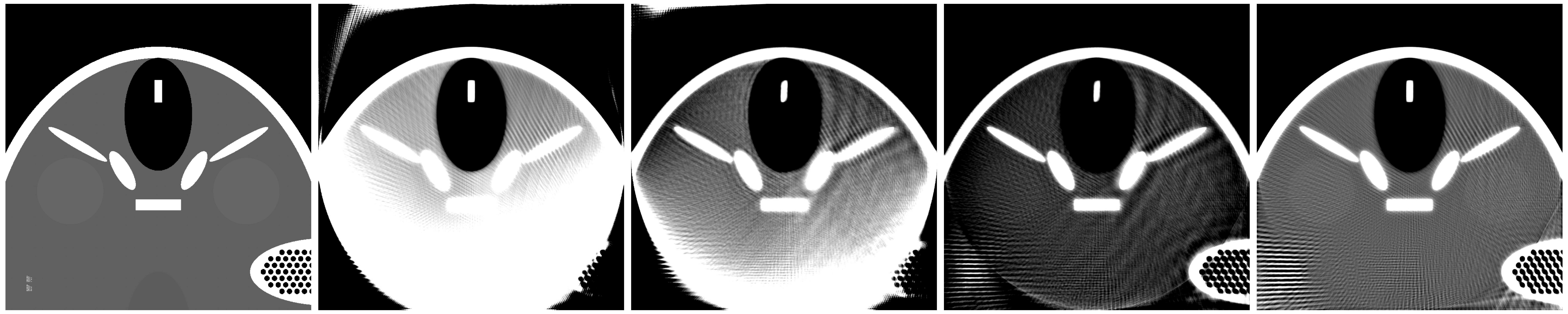}
\includegraphics[width=1.0\textwidth]{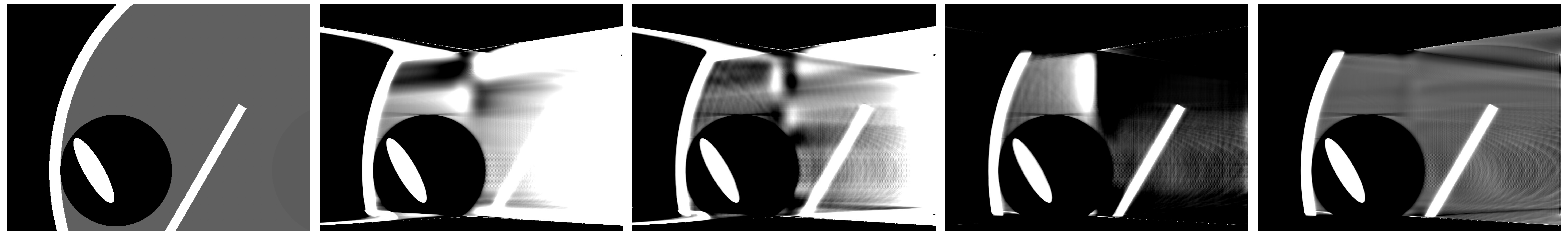}
\caption{Comparison of different reconstruction methods for the Forbild head phantom in 3D dental CBCT imaging. The first column represents the ground truth. The second column shows the results of the iterative method without a coarse grid prior. The third to fifth columns show the results of the iterative method with coarse grid reconstruction priors generated using FDK, iterative approach, and INR (proposed), respectively. The first and second rows correspond to axial and sagittal views, respectively. All CBCT images are displayed with the same window level and width.}
\label{result_comparison}
\end{figure*}

To evaluate the performance of the proposed algorithm, we compared the proposed method with different reconstruction approaches for the Forbild head phantom. All CBCT images were reconstructed under the dental CBCT geometry settings stated in Table \ref{tab_ct_geometry} and visualized on the fine grid detailed in Table \ref{tab_imaging_domain}. As shown in Figure \ref{result_comparison}, direct iterative reconstruction without a coarse grid prior suffers from severe bright shadowing artifacts due to FOV truncation (second column). While the iterative reconstruction methods with FDK-based \cite{FELDKAMP1984} and iterative-based coarse grid priors show some improvements, they still exhibit artifacts and contrast degradation (third and fourth columns). In contrast, the proposed method with the INR-based prior effectively reduces truncation artifacts and enhances the visibility of low-contrast objects (fifth column).

\section{Discussion and Conclusion}

In this study, we introduced a two-stage iterative reconstruction approach for dental CBCT with a truncated FOV. The INR-based prior, reconstructed at a coarse resolution over an extended FOV that fully encompasses the patient's head, enables accurate estimation of discrepancies in the projection data, which are then used to refine the projections before applying conventional iterative reconstruction within the truncated region. 

The experimental results using the Forbild head phantom demonstrated that direct iterative reconstruction without a coarse grid prior suffers from severe bright shadowing artifacts, which are particularly pronounced due to FOV truncation. The INR-based approach effectively captures complex spatial relationships even in the presence of missing projection data caused by the small detector. Incorporating INR-based coarse grid priors into the iterative reconstruction framework successfully suppressed truncation artifacts and improved the visibility of low-contrast objects, as observed in the axial view of Figure \ref{result_comparison}.

While the proposed approach demonstrated notable improvements in truncation artifact reduction, certain limitations remain. First, the INR reconstruction process introduces additional computational steps compared to direct conventional iterative methods, increasing overall processing time and memory burden, particularly in high-resolution 3D CBCT applications. Second, the proposed method was evaluated using a phantom-based experiment, and its performance in a real clinical environment with patient data remains to be validated, and this will be addressed in future work. 

\section*{References}
%\bibliographystyle{medphy.bst}    %if this is installed on your system,
%\bibliography{references}     %example.bib is on the same directory

\end{document}